# Evaluating Cross-region Generalization for Wavelet-Diffusion Precipitation Downscaling

**Weikang Qian[1], Yixin Wen[2,3]†, Chugang Yi[4], Zhi Li[5,6], Lingcheng Li[7], and Haizhao Yang[4,8]**

[1]University of Florida, Department of Geography, Gainesville, FL, USA
[2]School of Marine and Atmospheric Sciences, Stony Brook University, Stony Brook, NY, USA
[3]Institute for Advanced Computational Science, Stony Brook University, NY, USA
[4]Department of Mathematics, University of Maryland, College Park, MD, USA
[5]Civil, Environmental, and Architectural Engineering, University of Colorado Boulder, Boulder, CO, USA
[6]Institute of Arctic and Alpine Research, University of Colorado Boulder, Boulder, CO, USA
[7]Atmospheric, Climate, and Earth Sciences Division, Pacific Northwest National Laboratory, Richland, WA, USA
[8]Department of Computer Science, University of Maryland, College Park, MD, USA
†Corresponding author: Yixin Wen (yixin.wen@stonybrook.edu)

## Abstract

Diffusion models have shown strong potential for kilometer-scale precipitation downscaling, but their performance in geographically unseen regions and event regimes remains insufficiently understood. Building on the wavelet diffusion model (WDM) framework, this study evaluates cross-region and cross-event generalization. Six 3° × 3° U.S. regions represent convective, winter, tropical, and atmospheric-river precipitation regimes. Low-resolution inputs are generated by block averaging NOAA Multi-Radar/Multi-Sensor (MRMS) composite reflectivity fields. A WDM trained only on Oklahoma (OK) samples and a WDM trained on all six regions are compared with nearest-neighbor and Bicubic interpolation.Model performance is evaluated using three metric families that measure image-domain reconstruction, spectral and distributional fidelity, and bin-wise precipitation detection. The OK-trained WDM remains competitive outside OK. Although the all-region WDM delivers the best and most consistent overall image-domain and detection performance, its gains are uneven across precipitation intensities. Bin-wise critical success index (CSI) over 5-dBZ reflectivity bins shows that WDM improvements concentrate in localized higher-reflectivity structures, which image-domain metrics partly obscure. In addition, the performance differences among samples are strongly associated with the spatial organization of the precipitation field, quantified by Moran's I as the spatial autocorrelation of each reflectivity bin. The sample-level Moran's I-CSI correlation stratified by sample intensity reaches 0.901 in all six regions, including regions unseen during training. This indicates that the spatial organization contributes more precipitation downscaling tasks than are region and event difference. Overall, these findings support future efforts to transfer downscaling models to regions with limited local training data and to generate globally consistent, high-resolution precipitation products.

# 1 Introduction

High-resolution precipitation fields are essential for local hydrological modeling, flood-risk assessment, and hazard management, but many gridded precipitation products and model outputs are too coarse to resolve convective cells, localized heavy-rainfall cores, and sub-grid spatial variability (Li et al., 2022; Peleg et al., 2018; Prein et al., 2015; Sun et al., 2018; Wen et al., 2021). In addition to coarse nominal resolution, remotely sensed precipitation products are affected by spatial and temporal sampling errors, further complicating their use in fine-scale hydrologic and hazard applications (Behrangi & Wen, 2017). Deep-learning models have become a practical approach for precipitation downscaling, with methods evolving from early CNN-based downscaling to recent generative models that better represent fine-scale variability and heavy-precipitation structures (Harris et al., 2022; Leinonen et al., 2021; Ling et al., 2024; Liu et al., 2025; Mardani et al., 2025; Ravuri et al., 2021; Rodrigues et al., 2018; Vandal et al., 2017). However, most of these advances have been evaluated primarily using within-region validation, leaving their cross-region generalization less well understood.

Strong performance under within-region validation does not necessarily imply reliable performance in geographically unseen regions. This is a general problem for machine-learning systems under distribution shift, where models can perform well within the training distribution but degrade when deployed in a different domain (Koh et al., 2021; Liu et al., 2023). The out-of-distribution (OOD) problem is especially relevant for precipitation downscaling because regional precipitation regimes differ in their intensity distribution, terrain influence, storm structure, and spatial dependence. A model trained in one region may therefore learn mappings that are locally effective but less transferable to another region. Only a few studies have begun to examine this issue in precipitation downscaling, typically through isolated case assessments rather than systematic cross-region evaluation (Glawion et al., 2025). RainShift formalized precipitation downscaling under geographic distribution shifts and showed that out-of-distribution regions can produce substantial performance losses, while broader training domains do not always remove the gap (Harder et al., 2025). PrecipDiff further illustrates the need for caution when extending diffusion-based precipitation enhancement models beyond their evaluation domain (Dai & Ushijima-Mwesigwa, 2025). Thus, a focused evaluation of cross-region transfer and multi-region training is needed for generative precipitation downscaling.

This generalization problem also requires precipitation-structure-aware evaluation. Pixel-wise metrics and image similarity scores are useful summaries, but they are incomplete for precipitation fields. Small spatial shifts in a realistically reconstructed storm core can be penalized strongly, while overly smooth predictions may still receive favorable aggregate scores. Spatial and neighborhood-based verification methods were developed partly to address this problem by evaluating precipitation structure over spatial tolerances rather than only point-to-point agreement (Ebert, 2008; Gilleland et al., 2009; Roberts & Lean, 2008). Recent work has continued to refine these ideas for high-resolution and global precipitation forecasts, including improved interpretation of neighborhood skill, scalable spatial verification, and threshold-weighted evaluation for extreme precipitation (Loveday &

Hertneky, 2026; Skok & Lledó, 2025). Thus, evaluating cross-region precipitation downscaling should extend beyond conventional statistical metrics to determine whether fine-scale precipitation structures are preserved across regions and to help explain sample-level variations in model performance.

Our previously developed wavelet diffusion model (WDM) (Yi et al., 2026) was designed for efficient kilometer-scale precipitation downscaling by modeling high-frequency wavelet coefficients within a conditional diffusion framework. It showed that WDM achieves stronger reconstruction performance and faster inference than U-Net-based baselines and alternative diffusion variants. That evaluation, however, was conducted within a single region, leaving open whether these advantages persist when the model is applied to unseen regions and event conditions. Building on this established architecture, the present study therefore keeps the model weights fixed and shifts the focus from architecture development to assessing cross-region model generalization, as well as identifying sample characteristics associated with model performance.

To answer these questions, we used an expanded MRMS-based dataset covering six heterogeneous precipitation regimes across the continental United States. We evaluated WDM under both region-specific and multi-region training settings and compared its performance across accuracy, spectral and distribution fidelity, intensity-stratified detection skill, and spatial structure diagnostics. This design addresses three objectives:

- to test whether a model trained in one region can transfer to geographically unseen regions,
- to determine whether multi-region training improves robustness relative to region-specific training, and
- to identify possible spatial characteristics associated with model performance.

Section 2 describes the MRMS dataset, synthetic coarse-to-fine sample construction, WDM configuration, and evaluation framework. Section 3 reports the cross-region and multi-region downscaling results. Section 4 interprets these results through event-level diagnostic analysis and discusses the operational boundaries of generative precipitation downscaling. Section 5 summarizes the main conclusions and future directions.

## 2 Data and Methods

### 2.1 MRMS dataset

MRMS integrates multiple radar and auxiliary data streams to support severe-weather, hydrological, aviation and numerical weather prediction applications (Smith et al., 2016; Zhang et al., 2016). Following WDM framework of Yi et al. (2026), this study uses composite reflectivity (CR) as a high-resolution representation of precipitation structure rather than quantitative precipitation estimation, thereby avoiding precipitation-rate retrieval uncertainty and focusing on radar-reflectivity morphology (Krajewski et al., 2010).

Six fixed regions (Figure 1a) were extracted from the continental United States: Oklahoma (OK), Illinois (IL), Washington (WA), Florida (FL), Maryland (MD), and California (CA). Each region spans approximately 3° × 3° and is cropped to a 300 × 300 high-resolution grid. The regions were selected to expose the model to different precipitation regimes rather than to create a single climatologically homogeneous benchmark. The events include severe convection (OK), winter precipitation (IL), heavy snow (WA), tropical-cyclone-related precipitation (FL), ice-storm and winter events (MD), and atmospheric-river-related precipitation (CA). Event time windows were obtained from the NOAA Storm Events Database for 2022-2024. For each region, we filtered records by event type or narrative. We then merged overlapping or adjacent intervals that referred to the same event type and converted local times to UTC using the recorded time-zone field. The event filters and spatial regions are summarized in Supporting Information (SI) Table S1.

### 2.2 High- and low-resolution sample construction

For each event interval, MRMS files were sampled every 10 min from the public noaa-mrms-pds archive. Each file was cropped to the corresponding region. Missing values marked as -99 and -999 were set to zero, and a time slice was retained only if more than 10% of the grid cells had reflectivity greater than zero (roughly 0.027 mm/hr rainfall if applying the standard Marshall-Palmer relationship). This filtering step removed mostly empty scenes and concentrated the benchmark on samples containing precipitation. Each retained time slice is one sample.

The high-resolution target is the 300 × 300 cropped CR field. The low-resolution input is generated by 10 × 10 block averaging of the high-resolution field, giving a 30 × 30 grid. This design creates a controlled tenfold spatial downscaling problem in which the low-resolution and high-resolution fields share the same observing product and timestamp. We refer to this as the same-product setting. All models were trained on samples from 2022 and 2023 and tested on independent 2024 samples. The 2022-2023 subset was randomly split into 80% training and 20% validation samples. Supplementary Table S2 reports the resulting sample counts.

To describe the intensity regime of each sample while avoiding isolated-pixel maxima, we used a filtered maximum reflectivity bin, denoted Bmax25. Bmax25 is the highest 5-dBZ bin in the high-resolution target that contains at least 25 pixels. The threshold prevents a single isolated pixel from defining the sample maximum (at approximately 1-km grid spacing, 25 pixels correspond to an area on the order of 25 $km^2$). For example, a sample with Bmax25 = 50 dBZ can still contain lower bins from 0 to 50 dBZ, but it may not contain the 55-dBZ bin or may contain it with fewer than 25 pixels. Bmax25 is therefore treated as a robust sample-level intensity characteristic. Figure 1b shows the Bmax25 distribution in the 2024 test set. The distribution shows that the test set covers a range of event intensities and that the intensity distribution differs substantially across regions.

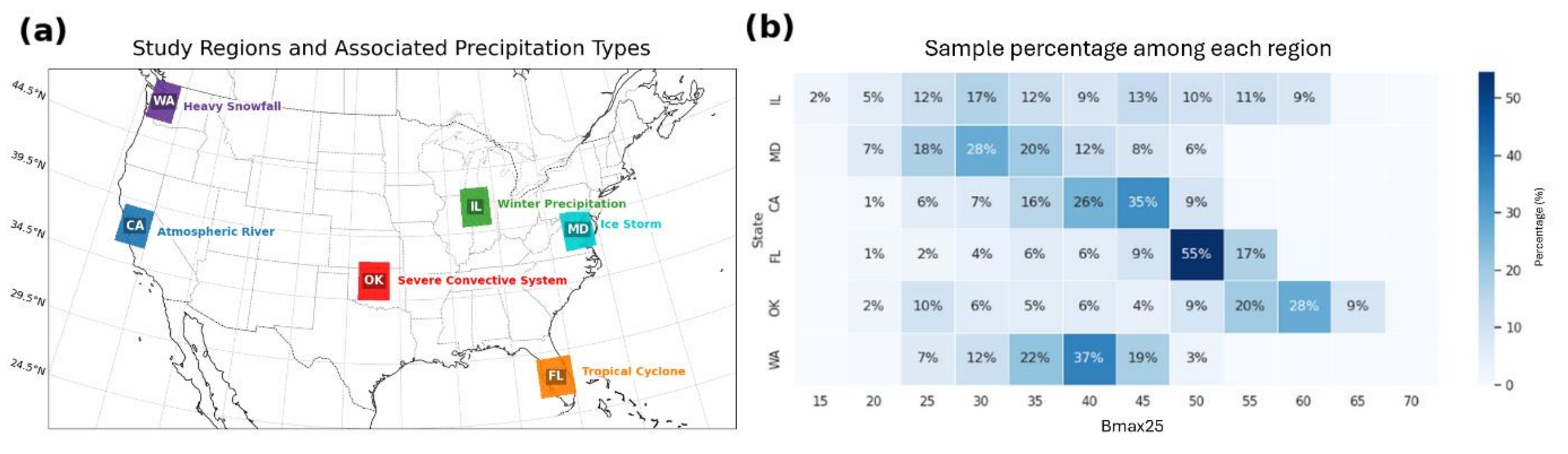

**Figure 1. Study domain and test-set regime distribution. (a) Six 3° × 3° MRMS crop regions; Oklahoma is highlighted because it is the single-region training domain for WDM-OK model. (b) Distribution of 2024 test samples by region and support-filtered maximum reflectivity bin Bmax25. Percentages are computed within each region.**

## 2.3 Wavelet diffusion model (WDM) and interpolation baselines

The target framework is WDM, a conditional wavelet diffusion model for precipitation downscaling (Jiang et al., 2023; Phung et al., 2023; Yi et al., 2026). The model conditions the generation process on the low-resolution input and performs diffusion-based downscaling in a wavelet representation through a discrete wavelet transformation of the original CR field. The wavelet representation is useful for this task because the missing information between 30 × 30 and 300 × 300 fields is dominated by high-frequency spatial detail, while the coarse input already constrains broad event position and intensity. A single-level discrete wavelet transform (DWT) was used in this study.

Two WDM variants are evaluated. WDM-OK model (WDM-OK) is trained only on the OK training set, making it a single-region model tested both within distribution and OOD. The reason for training in OK is that OK has the broadest spectrum of precipitation regimes, as shown in Figure 1b, with Bmax25 spanning across different reflectivity bins. WDM-ALL model (WDM-ALL) is trained on pooled training samples from all six regions and event types, making it a multi-region model. Both variants were trained in two target configurations: the original reflectivity field and a residual field. In the residual configuration, the model learns the difference between the high-resolution target and the upsampled low-resolution field in the wavelet domain, so that the generative component focuses on the missing fine-scale structure. Both WDM-OK and WDM-ALL models were trained for 5,000 epochs. Validation was performed every 20 epochs, and the checkpoint with the lowest validation mean-square error (MSE) was selected to avoid overfitting. Validation and inference used 300 denoising steps. The residual-field models were selected for the main evaluation because they converged faster and had comparable performance on the validation set, with 1-4% higher MSE than those of the original-field models. Additional training details are provided in SI Figure S1.

Nearest-neighbor (NN) and Bicubic are included as benchmark methods. NN upsampling expands the 30 × 30 low-resolution grid to 300 × 300 without introducing new spatial information. It therefore serves as a low-resolution proxy and a minimum standard that learned models should exceed. Bicubic interpolation provides a classical, nonlearned smoothing baseline for precipitation downscaling. Other AI or generative baselines are not retrained here because the original WDM study compared WDM with broader model families, including EDM and FourierDM (Karras et al., 2022; Yi et al., 2026).

## 2.4 Evaluation metrics

### 2.4.1 General downscaling metrics

Peak signal-to-noise ratio (PSNR) and structural similarity index measure (SSIM) were used to assess general image quality. PSNR measures the magnitude of pixel-level reconstruction error relative to the dynamic range of the field, whereas SSIM evaluates similarity in local luminance, contrast and structural patterns (Wang et al., 2004). Coefficient of determination ($R^2$) and root-mean-square error (RMSE) were used to quantify pointwise downscaling accuracy. $R^2$ measures the fraction of target-field variance explained by the downscaled field, while RMSE reports the average downscaling error in the same units as the evaluated reflectivity arrays.

Beyond image quality and pointwise accuracy, spectral and distributional fidelity were also evaluated. Fourier-amplitude MAE compares the magnitude spectra of the predicted and target fields, providing a measure of whether the downscaling preserves the correct distribution of spatial scales. One-dimensional (1-D) Wasserstein distance compares the one-dimensional empirical pixel-value distributions of the prediction and target, allowing us to assess similarity in the intensity-distribution space. The mathematical definitions of these metrics are provided in SI Text S1.

### 2.4.2 Bin-wise detection metrics and improvement cases

Detection metrics are important in precipitation verification because warnings depend on correctly identifying relevant reflectivity features. In this context, fractions skill score (FSS) and critical success index (CSI) are commonly used for precipitation detection and high-resolution verification (Ebert, 2008; Gilleland et al., 2009; Roberts & Lean, 2008; Schaefer, 1990). FSS emphasizes detection skill within a neighborhood or coverage scale, whereas CSI is a stricter categorical score based on hits, false alarms and misses. Both are threshold-based metrics, so the selected threshold determines which precipitation structure is being evaluated. Empirically, reflectivity thresholds of 20 and 40 dBZ are used for precipitation and heavy-precipitation cases, respectively. To capture the ability of WDM to reconstruct both low and heavy precipitation, Yi et al. (2026) calculated CSI averaged the thresholds of [0, 20, 40, 50, 65] dBZ as one of the general metrics. Extending this multi-intensity averaging, we examined detection performance in greater detail. Rather than selecting thresholds empirically, we calculated CSI for each 5-dBZ bin from 0 to 70 dBZ to evaluate whether particular intensity ranges are detected, denoted bin-wise CSI. Rather than using a single threshold to create the binary mask, we applied both upper and lower bounds, investigating detection performance for each desired reflectivity range. For each bin, the target and prediction are converted into binary masks indicating whether a pixel falls within that bin. For a 5-dBZ interval beginning at $\tau$, where $\tau \in [\,0,5,\dots,70]$, we count pixels classified as true-positive (TP), false-positive (FP) and false-negative (FN).

Model improvements are categorized at the sample-bin level relative to a baseline. Let $\Delta CSI = CSI_{model} - CSI_{baseline}$ and $\Delta I = \frac{\Delta CSI}{CSI_{baseline}} * 100\%$. A positive improvement occurs when $\Delta CSI > 0$. A significant improvement occurs when the model detects a bin that the baseline misses, or when the gain exceeds 100%. This categorization separates ordinary gains from cases where the baseline essentially fails, which is important in high-reflectivity bins with limited pixels. Thus, the samples are classified into three categories: negative, normal-improvement and significant-improvement cases.

**2.4.3 Spatial autocorrelation diagnostics**

To diagnose why some events are easier to downscale than others, we computed Moran's I on the high-resolution target field at each 5-dBZ bin. Moran's I measures spatial autocorrelation (Moran, 1950). Throughout this paper, spatial autocorrelation, quantified by Moran's I, serves as a proxy of spatial organization. For a binary bin-wise field x, Moran's I can be written as

$$I = \frac{n}{S_0} \frac{\sum_{i=1}^{n} \sum_{j=1}^{n} \omega_{ij} \, z_{x,i} z_{y,j}}{\sum_{i=1}^{n} z_{x,i}^2}$$

where $z_{x,i} = \frac{x_i - \bar{x}}{s_x}$, $z_{y,j} = \frac{y_i - \bar{y}}{s_y}$, $S_0 = \sum_{i=1}^{n} \sum_{j=1}^{n} \omega_{ij}$. The spatial weights $\omega_{ij}$ are defined by queen contiguity and row-standardization on a regular grid. $s_x$ and $s_y$ are the standard deviations of x and y, respectively. Moran's I is used as a spatial autocorrelation diagnostic rather than as a direct performance metric. We correlate bin-wise Moran's I with bin-wise CSI and also compute sample-level weighted averages over valid bins. These diagnostics test whether reconstruction skill is conditioned by event morphology and spatial precipitation patterns rather than only by region label or maximum intensity. Moran's I typically ranges from -1 to 1, where positive values indicate positive spatial autocorrelation, meaning that pixels with similar bin membership tend to be spatially clustered; values close to zero indicate little spatial organization; and negative values indicate spatial dispersion or checkerboard-like patterns.

# 3 Results

## 3.1 General downscaling metrics

To test whether WDM-OK model transfers to unseen regions, Table 1 compares it against Bicubic interpolation across all six regions, showing that its OOD performance depends strongly on the metric family. Broadly, WDM-OK has the lowest 1D Wasserstein distance, which measures the probability distribution of radar reflectivity in the frequency domain. However, Bicubic remains slightly better for the image-domain metrics (PSNR, $R^2$ and RMSE) in the four remaining external regions (IL, MD, CA, and WA), indicating that WDM-OK is competitive but not uniformly superior outside the source region. In IL, for example, amplitude MAE and Wasserstein distance decrease from 1.52 and 0.54 for Bicubic to 1.22 and 0.36 for WDM-OK, respectively. These results indicate that WDM-OK better preserves the spectral and distributional properties of precipitation fields, even outside its source region.

WDM-ALL achieved the strongest mean image-domain performance. It had the highest PSNR and the lowest RMSE in all six regions, increasing PSNR by 1.14–1.79 dB and reducing RMSE by 12.6–18.0% relative to NN. The relative RMSE reduction is largest in MD (from 2.22 to 1.82) and smallest in CA (from 3.80 to 3.32). WDM-ALL also performs slightly better than WDM-OK model in the OK region for the image-domain metrics, indicating that adding samples from other regions does not weaken image-domain performance in the source region. However, WDM-OK still outperforms WDM-ALL in amplitude MAE and Wasserstein distance across all six regions. Notably, WDM-ALL does not outperform NN in amplitude MAE in four of the six regions, suggesting that pooled training may trade amplitude MAE fidelity for accuracy in the image-domain metrics. Figure 2 summarizes these differences by expressing each metric as improvement relative to NN. The radar plots confirm the pattern in Table 1 across all six regions. WDM-ALL leads in the image-domain metrics, while WDM-OK retains its advantage in amplitude MAE and Wasserstein distance.

**Table 1. Reconstruction metrics on the 2024 test set. Higher values are better for PSNR, SSIM and $R^2$; lower values are better for RMSE, amplitude MAE and Wasserstein distance. Values are rounded to two decimals.**

| Region | PSNR↑ | | | | SSIM↑ | | | | $R^2$↑ | | | |
|---|---|---|---|---|---|---|---|---|---|---|---|---|
| | ALL | OK | Bicubic | NN | ALL | OK | Bicubic | NN | ALL | OK | Bicubic | NN |
| **OK** | ***29.90*** | 29.60 | 29.47 | 28.47 | ***0.59*** | 0.58 | 0.58 | 0.54 | ***0.86*** | 0.85 | 0.85 | 0.82 |
| **IL** | ***31.11*** | 30.59 | 30.74 | 29.72 | ***0.60*** | 0.58 | 0.60 | 0.55 | ***0.88*** | 0.86 | 0.87 | 0.84 |
| **MD** | ***33.22*** | 32.49 | 32.72 | 31.43 | ***0.65*** | 0.63 | 0.65 | 0.59 | ***0.91*** | 0.89 | 0.90 | 0.87 |
| **FL** | ***27.52*** | 27.07 | 27.00 | 25.90 | ***0.61*** | 0.59 | 0.59 | 0.53 | ***0.86*** | 0.85 | 0.84 | 0.81 |
| **CA** | ***27.93*** | 27.42 | 27.51 | 26.79 | ***0.54*** | 0.52 | 0.51 | 0.48 | ***0.76*** | 0.73 | 0.74 | 0.70 |
| **WA** | ***28.53*** | 27.91 | 28.13 | 27.32 | ***0.54*** | 0.51 | 0.51 | 0.47 | ***0.80*** | 0.78 | 0.79 | 0.76 |

| Region | RMSE↓ | | | | Amplitude MAE↓ | | | | Wasserstein Distance↓ | | | |
|---|---|---|---|---|---|---|---|---|---|---|---|---|
| | ALL | OK | Bicubic | NN | ALL | OK | Bicubic | NN | ALL | OK | Bicubic | NN |
| **OK** | ***2.63*** | 2.72 | 2.76 | 3.13 | 1.66 | ***1.42*** | 1.79 | 1.64 | 0.44 | ***0.39*** | 0.58 | 0.53 |
| **IL** | ***2.35*** | 2.49 | 2.46 | 2.78 | 1.40 | ***1.22*** | 1.52 | 1.39 | 0.41 | ***0.36*** | 0.54 | 0.48 |
| **MD** | ***1.82*** | 1.97 | 1.92 | 2.22 | 1.07 | ***0.92*** | 1.17 | 1.08 | 0.27 | ***0.22*** | 0.37 | 0.32 |
| **FL** | ***3.48*** | 3.65 | 3.70 | 4.22 | 1.90 | ***1.64*** | 2.11 | 1.90 | 0.54 | ***0.50*** | 0.88 | 0.75 |
| **CA** | ***3.32*** | 3.51 | 3.48 | 3.80 | 2.00 | ***1.78*** | 2.21 | 1.97 | 0.71 | ***0.64*** | 0.99 | 0.88 |
| **WA** | ***3.07*** | 3.29 | 3.22 | 3.54 | 1.86 | ***1.64*** | 2.04 | 1.84 | 0.64 | ***0.58*** | 0.89 | 0.79 |

Taken together, Table 1 and Figure 2 show that general metrics are necessary but not sufficient for evaluating precipitation downscaling, since they expose conflicts among image-domain, spectral, and distributional criteria. These conflicts highlight the limitations of traditional image-domain metrics for precipitation downscaling. The following section therefore turns to bin-wise CSI analysis, which evaluates detection skill directly across reflectivity ranges rather than inferring it from a small set of global image-domain metrics.

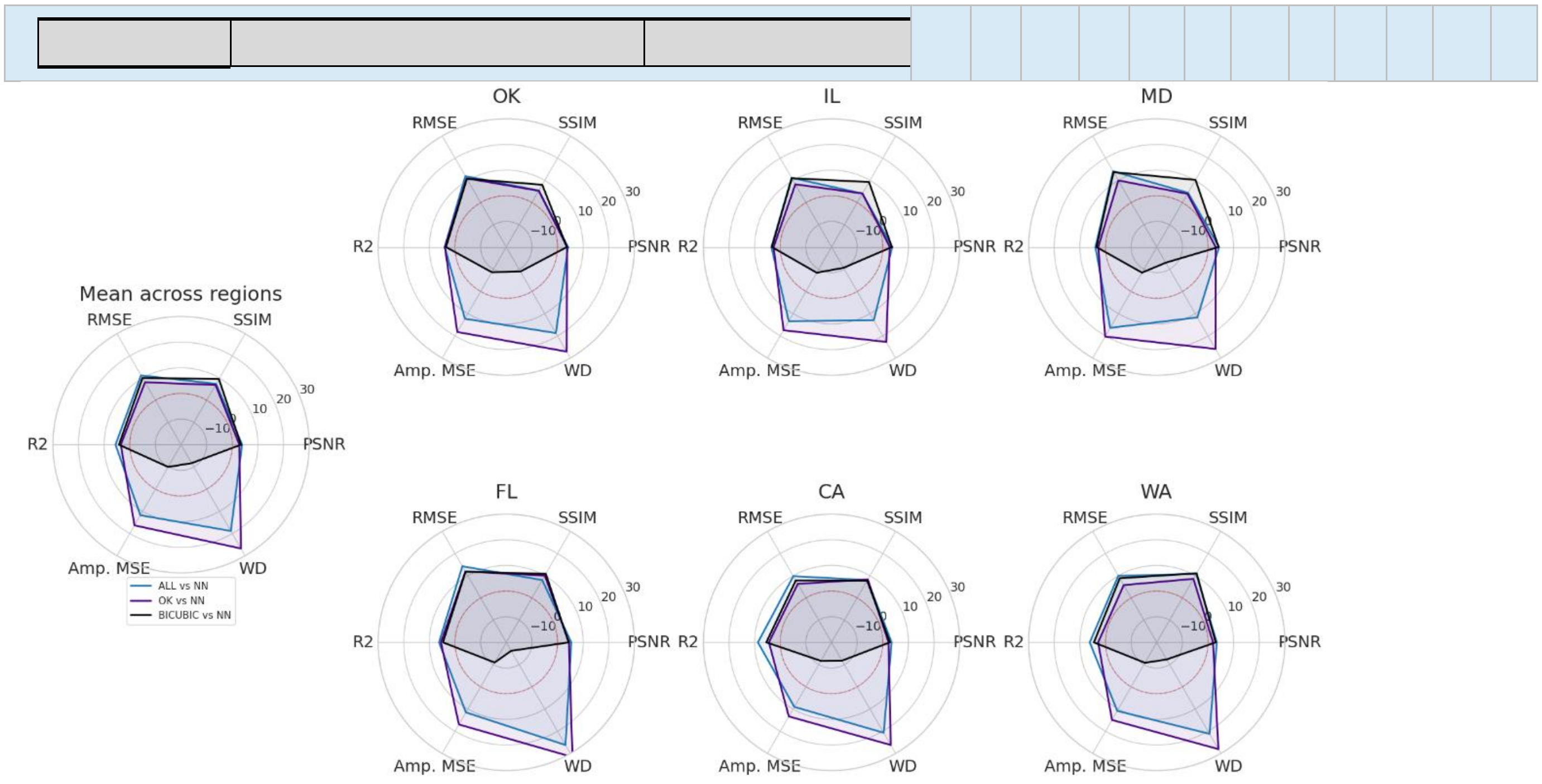

**Figure 2. Improvement relative to the NN baseline for general metrics. For metrics in which lower values are better, improvement is defined as reduction relative to NN; for metrics in which higher values are better, improvement is defined as gain relative to NN.**

## 3.2 Bin-wise CSI for precipitation detection

The general downscaling metrics do not provide a fully consistent basis for evaluation, particularly in OOD settings, because they emphasize different aspects of reconstruction quality. Thus, we used bin-wise CSI to evaluate the model from a precipitation detection perspective and to examine whether its performance is retained across different reflectivity ranges. Figure 3 compares bin-wise CSI for WDM-OK relative to Bicubic after pooling all six regions. Each 5-dBZ bin is summarized by normal-improvement and significant-improvement categories. The two categories represent different detection regimes. Normal-improvement cases generally have higher CSI means, showing that WDM adds skill when the target bin remains partly recoverable by Bicubic. Significant cases usually have lower CSI means, but they are diagnostically important because Bicubic misses the bin or provides only negligible CSI. The labels above the yellow boxes show the mean positive CSI improvement for normal-improvement cases. These means generally increase with reflectivity range, rising from 4.5-8.3% at 5-15 dBZ to 41.4-69.6% at 45-55 dBZ and above 90% at 60-65 dBZ. This trend shows that, even outside the significant-improvement category, WDM contributes more as reflectivity structures become harder for the baseline to preserve.

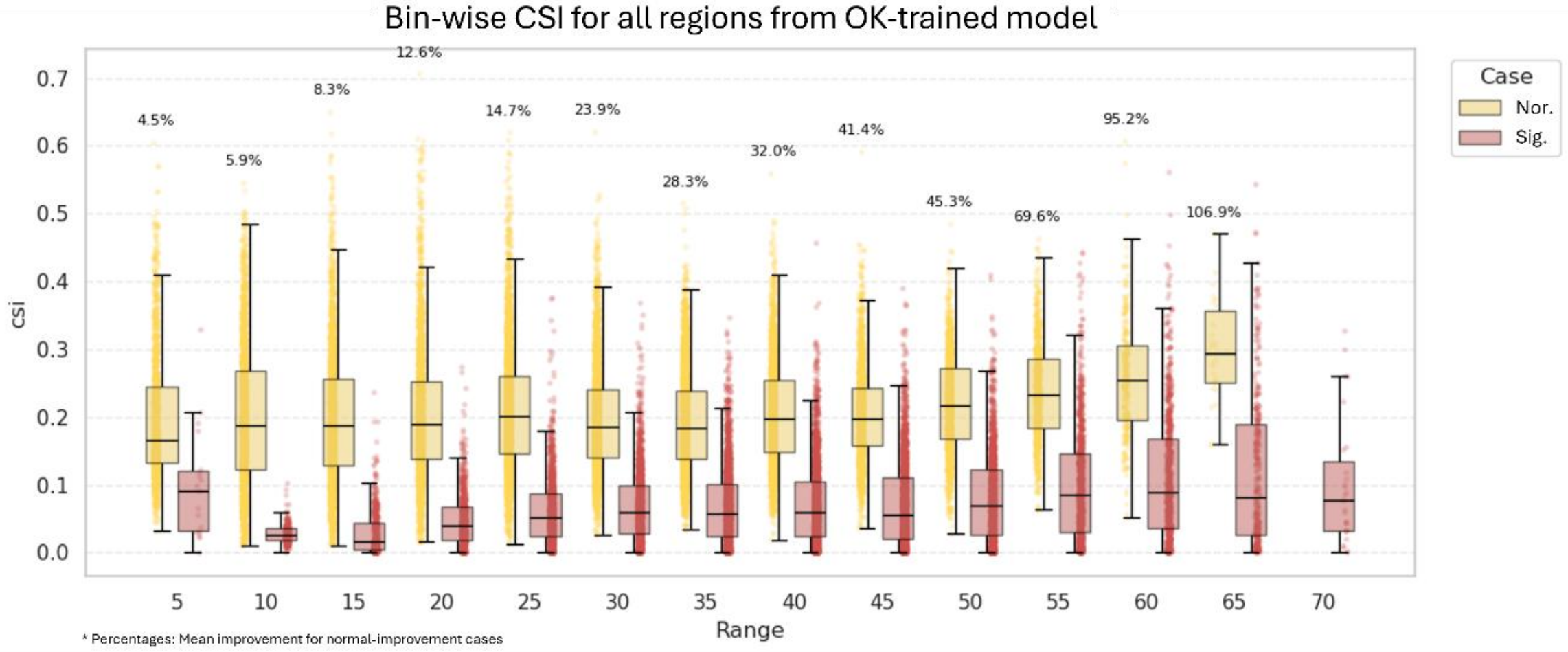

**Figure 3. Bin-wise CSI improvement of WDM-OK over Bicubic across all regions. Yellow boxes indicate normal-improvement cases; red boxes indicate significant-improvement cases. Numbers above the yellow boxes give the mean improvement for normal-improvement cases.**

Figure 4 extends the pooled comparison for normal cases by separating regions and comparing WDM-ALL, WDM-OK model and Bicubic against NN. Figure 4a reports the percentage of samples numbers with normal, significant, and negative improvement. Across all regions, the three methods show similar patterns in the normal-improvement cases. In low-reflectivity bins, the positive improvement fraction is high for all three models in most regions. This pattern shows that weak and broad reflectivity structures are often recoverable from the coarse input regardless of the applied method, and Bicubic sometimes outperforms both the WDM-ALL and WDM-OK models. As reflectivity increases, the normal improvement fraction declines.

The three models show different behavior in Figure 4a. Across 5-20 dBZ, normal-improvement cases averages are 91.7% for WDM-ALL, 81.3% for WDM-OK and 89.6% for Bicubic. This indicates that low-reflectivity structures are often recoverable from the coarse input, including by interpolation. However, the significant and negative cases show a clearer model difference. In low-reflectivity bins (<15 dBZ), negative cases account for 4.5% for WDM-ALL, compared with 14.9% for WDM-OK and 6.8% for Bicubic overall. Notably, the negative-case percentage for Bicubic peaks around 30 dBZ, while most negative-case percentages remain below 10% for WDM models. Thus, Bicubic remains competitive in low-reflectivity bins, but WDM-ALL has the most stable improvement-category profile. At higher reflectivity, the percentage of significant-improvement cases increases. For example, across 40-55 dBZ, the significant-improvement category averages 26.0% for WDM-ALL and 25.0% for WDM-OK, but only 3.0% for Bicubic. This pattern shows that WDM more often recovers high-reflectivity bins that Bicubic fails to detect. Combining normal and significant cases, WDM-OK improves in 73.8% of OK-region cases and 86.4% of external-region cases in the 5-20 dBZ range. In relatively high bins, the combined improvement percentage remains substantial both in OK and in the external regions. This indicates that WDM-OK's high-reflectivity detection behavior is not confined to the source region. Overall, Figure 4a indicates that all models perform similarly for weak reflectivity but diverge as reflectivity increases. The two WDM models, especially WDM-ALL, provide stronger and more stable improvement profiles by increasing significant gains in high-reflectivity bins while limiting negative cases. WDM-OK improves many external-region samples, although the higher external-region percentages may partly reflect differences in precipitation regimes across regions.

Figure 4b shows the mean CSI improvement including normal and limited negative cases. This separation distinguishes the frequency of improvement from the magnitude of improvement. WDM-ALL produces larger mean CSI gains than WDM-OK in all ranges. Across 5-20 dBZ, the mean improvement is 17.2% for WDM-ALL and 12.7% for WDM-OK. Across 40-55 dBZ, the corresponding values are 18.6% and 15.0%. Bicubic behaves differently. In the low-reflectivity range, its mean improvement is also 12.7%, close to WDM-OK. This low-bin competitiveness helps explain why Bicubic can remain competitive according to PSNR, SSIM, $R^2$ and RMSE. The high-reflectivity bins reveal the main limitation of Bicubic interpolation. Across 40-55 dBZ, its mean CSI improvement is only 0.3%, and negative mean improvement appears in this range. WDM mitigates the main interpolation failure mode: smoothing can preserve broad image similarity, which strongly influences aggregate image-domain metric, but it can also remove localized high-reflectivity features that matter for precipitation downscaling.

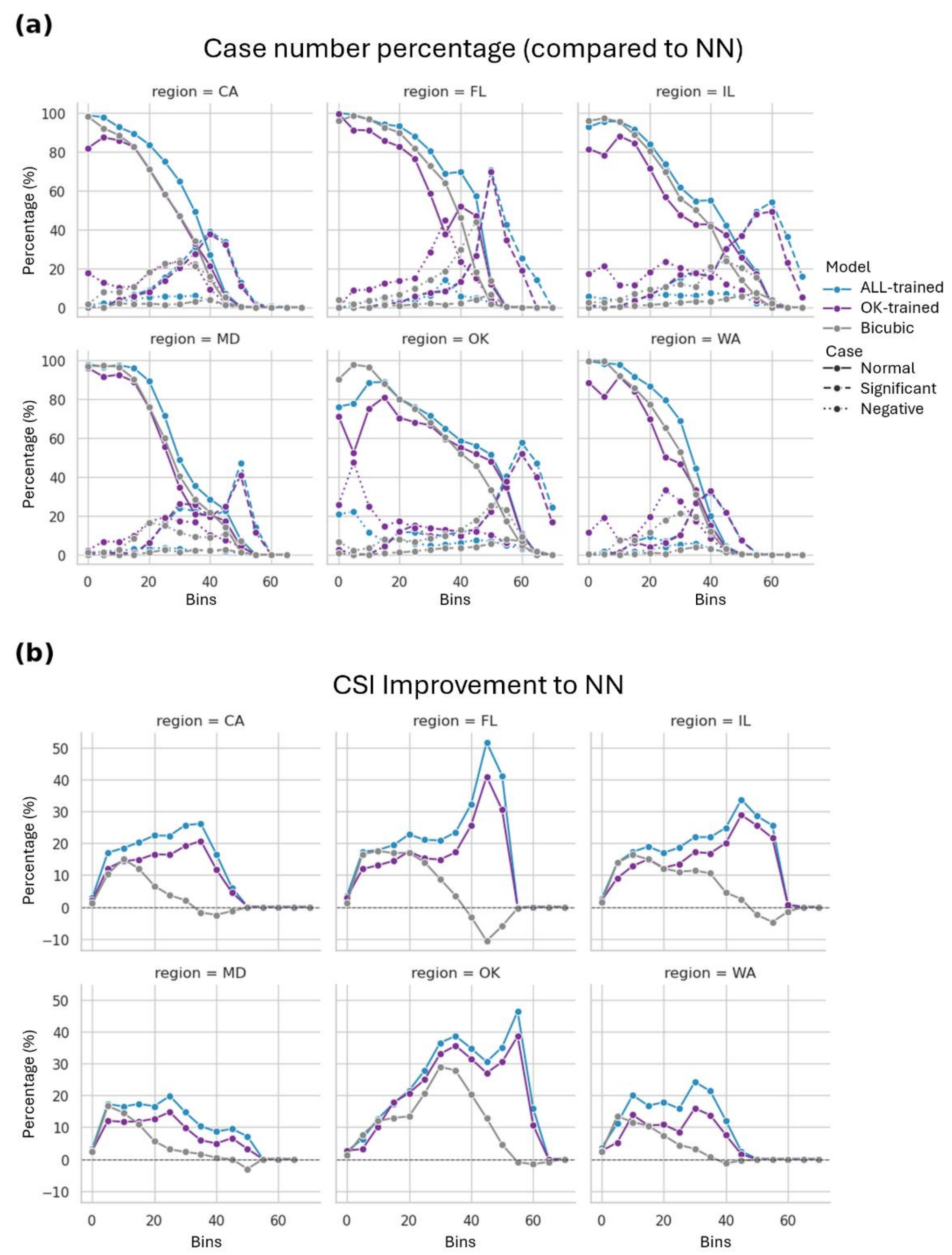

**Figure 4. CSI improvement patterns by region and reflectivity bin. (a) Percentages of sample-bin cases with normal, significant, and negative improvement relative to NN. (b) Mean normal CSI improvement by bin. Lines represent the evaluated methods: blue denotes WDM-ALL, purple denotes WDM-OK, and gray denotes bicubic interpolation.**

## 3.3 Case studies

Figure 5 provides representative cases from the six study regions to clarify what the quantitative metrics measure and where they disagree at the sample level. NN retains blocky 10-km-scale structures because it only expands the coarse grid. Bicubic smooths sharp gradients and tends to blur high-reflectivity cores. WDM outputs recover more spatially coherent fine-scale structures. The two WDM variants differ mainly in smoothness and local intensity placement. WDM-ALL generally appears smoother, whereas WDM-OK often retains sharper, finer-scale texture.

The ALL - OK panels show that these differences concentrate near precipitation boundaries, embedded cores and fragmented structures. The error maps also show remaining failure locations. Errors concentrate near sharp transitions, small isolated cells and intense cores whose exact location is difficult to infer from a 30 × 30 coarse input. A small displacement of an intense core may produce a large pixel-wise penalty even if the reconstruction is meteorologically plausible, while a smoothed core may keep RMSE moderate but fail bin-wise detection.

More specifically, the OK and FL examples illustrate why visual quality and general metrics do not always give the same ranking. In the OK case, which is within the source region, Bicubic achieves the highest SSIM (0.616), because it preserves the broad storm structure and suppresses small-scale variation. However, it visibly smooths the high-reflectivity band and does not reproduce the fine texture of the high-resolution target. WDM-OK output has the lowest Wasserstein distance (0.270), indicating a closer match to the target intensity distribution, while WDM-ALL gives the best PSNR and RMSE, suggesting better pixel-level agreement. In the FL case, which is an external-region example from a landfalling tropical cyclone, Bicubic again retains the large-scale precipitation pattern but broadens and smooths the organized rainband structure. Both WDM outputs better recover the fragmented and filamentary precipitation features, and WDM-OK gives the lowest Wasserstein distance. Thus, the case studies demonstrate the need for both image-domain and precipitation-specific metrics.

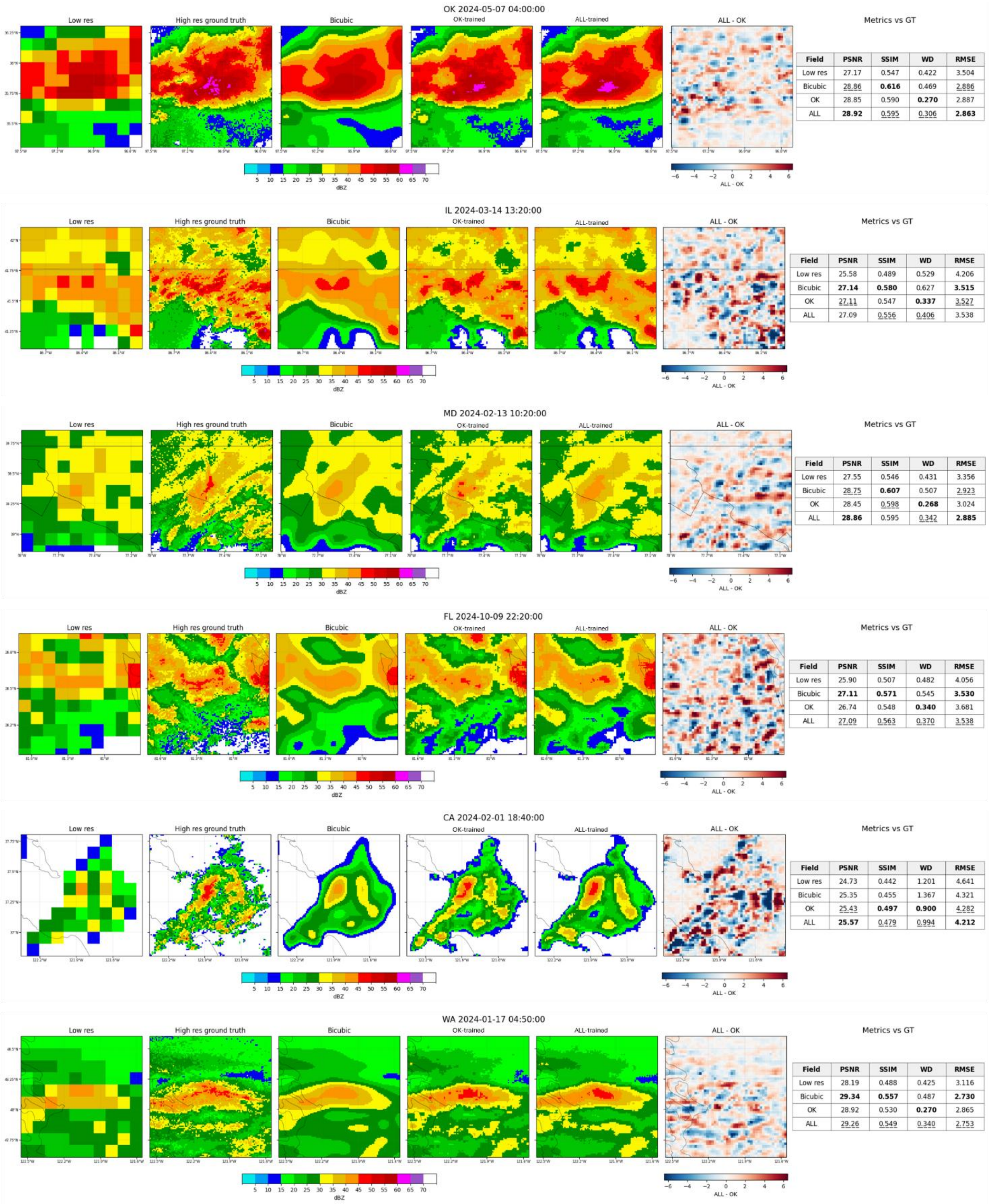

**Figure 5. Representative downscaling cases from the six regions, a cropped vision (100*100). Each row shows one 2024 sample. Columns show the low-resolution input, high-resolution ground truth (target), Bicubic, WDM-OK, WDM-ALL, ALL - OK difference field, and case-level metrics computed against the high-resolution target. The ALL - OK panel shows local differences between the two WDM variants.**

# 4 Discussion

The results demonstrate meaningful cross-region generalization, with WDM-OK remaining competitive in unseen regions and WDM-ALL delivering the most stable performance across regions. However, the results also reveal several issues that require further discussion. First, the large CSI variability among samples with comparable reflectivity or intensity conditions, as shown in Figure 3, indicates that intensity alone cannot explain downscaling performance. Second, the difference between image-domain and reflectivity-bin metrics shows the need for evaluation metrics that directly measure reconstruction of the spatial patterns. Third, the strong cross-region performance requires careful interpretation in relation to previous studies that report OOD degradation in precipitation downscaling. This section discusses these issues by analyzing the relationship between bin-wise CSI and Moran's I spatial autocorrelation, and by defining the task boundary of regional robustness.

## 4.1 Spatial autocorrelation helps explain within-bin variability in downscaling performance

Samples with comparable reflectivity can differ strongly in spatial organization. A 40-45 dBZ bin, for example, can appear as an isolated convective core, a quasi-linear system, or a cluster of storms. These patterns have similar intensities, but they leave different information in the coarse field. When target pixels are spatially clustered, the coarse field can retain useful information about their location and shape. These differences suggest that spatial organization is one source of within-bin CSI variability. To test this explanation, we use bin-wise Moran's I to measure whether pixels within a reflectivity bin are spatially clustered or scattered.

Figure 6 directly addresses the large CSI variability observed within individual reflectivity bins in Figure 3. Across most bins, Moran's I and CSI show a broadly positive relationship. This relationship is clearest in the lower and middle reflectivity ranges. The all-case correlation coefficient (CC) is about 0.92–0.95 for the 5–20 dBZ bins and about 0.78–0.90 for the 25–45 dBZ bins. Part of this relationship is expected from the spatial patterns because scattered pixels are unlikely to survive tenfold block averaging. In contrast, clustered pixels leave a stronger imprint in the coarse field.

Normal and significant cases differ mainly in how widely CSI is dispersed around this relationship. In both categories, high-CSI cases concentrate in the high-Moran's-I region. The relationship becomes weaker in the highest reflectivity bins. The all-case CC decreases to about 0.60 at 60 dBZ, 0.49 at 65 dBZ, and 0.18 at 70 dBZ. This weakening is related to the small number of target pixels in these bins. High-reflectivity bins often correspond to compact convective cores, so a small number of missed or displaced pixels can strongly change CSI. A spatially clustered core may still receive a low CSI if it is slightly displaced. This sensitivity makes the relationship less regular near 70 dBZ.

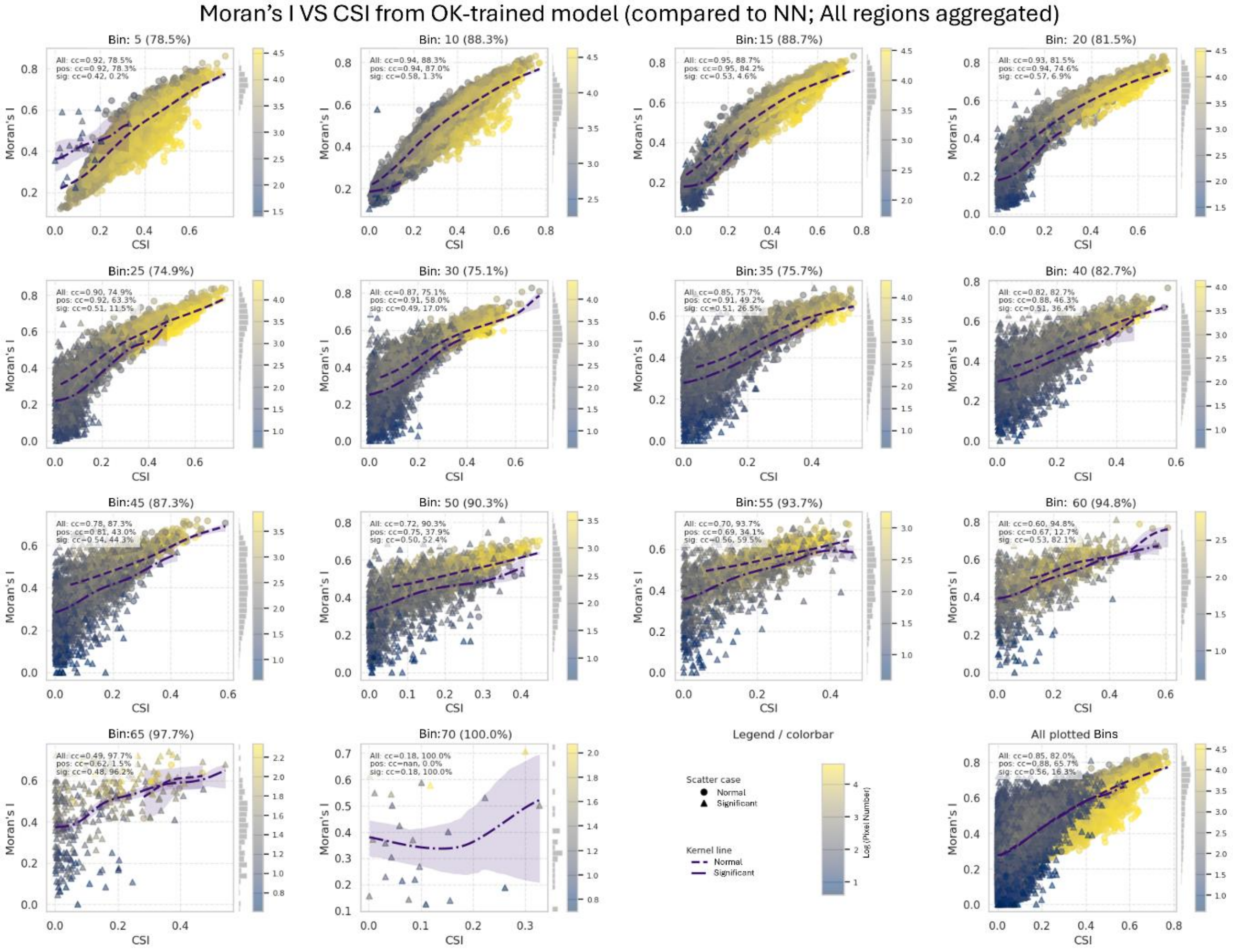

**Figure 6. Scatter plots of bin-wise Moran's I and CSI. Each panel corresponds to a reflectivity bin; the color scale indicates log pixel number, and percentages in panel titles summarize positive improvement cases to all cases.**

Figure 7 further summarizes the relationship between bin-wise Moran's I and CSI after stratifying the all cases by region and Bmax25. Bmax25 is used here as a sample-level intensity proxy, allowing the analysis to test whether the relationship persists across weaker and stronger precipitation samples. After this stratification, the correlations become stronger in most groups, commonly around 0.80 - 0.96. This suggests that mixing samples with different intensities and pixel numbers adds noise to the pooled relationship in Figure 6. Though there are still a limited categories with CC smaller 0.8, the consistently positive correlations across most region-Bmax25 groups show that the relationship is not limited to a specific region or intensity range. Together, Figure 6 and Figure 7 show that spatial autocorrelation, quantified by Moran's I, helps explain the CSI variability within comparable reflectivity conditions, with the strength of this relationship further shaped by pixel count and sample intensity. Away from the highest bins, high-reflectivity structures can still be recovered when their pixels show high spatial autocorrelation and sufficient pixel number.

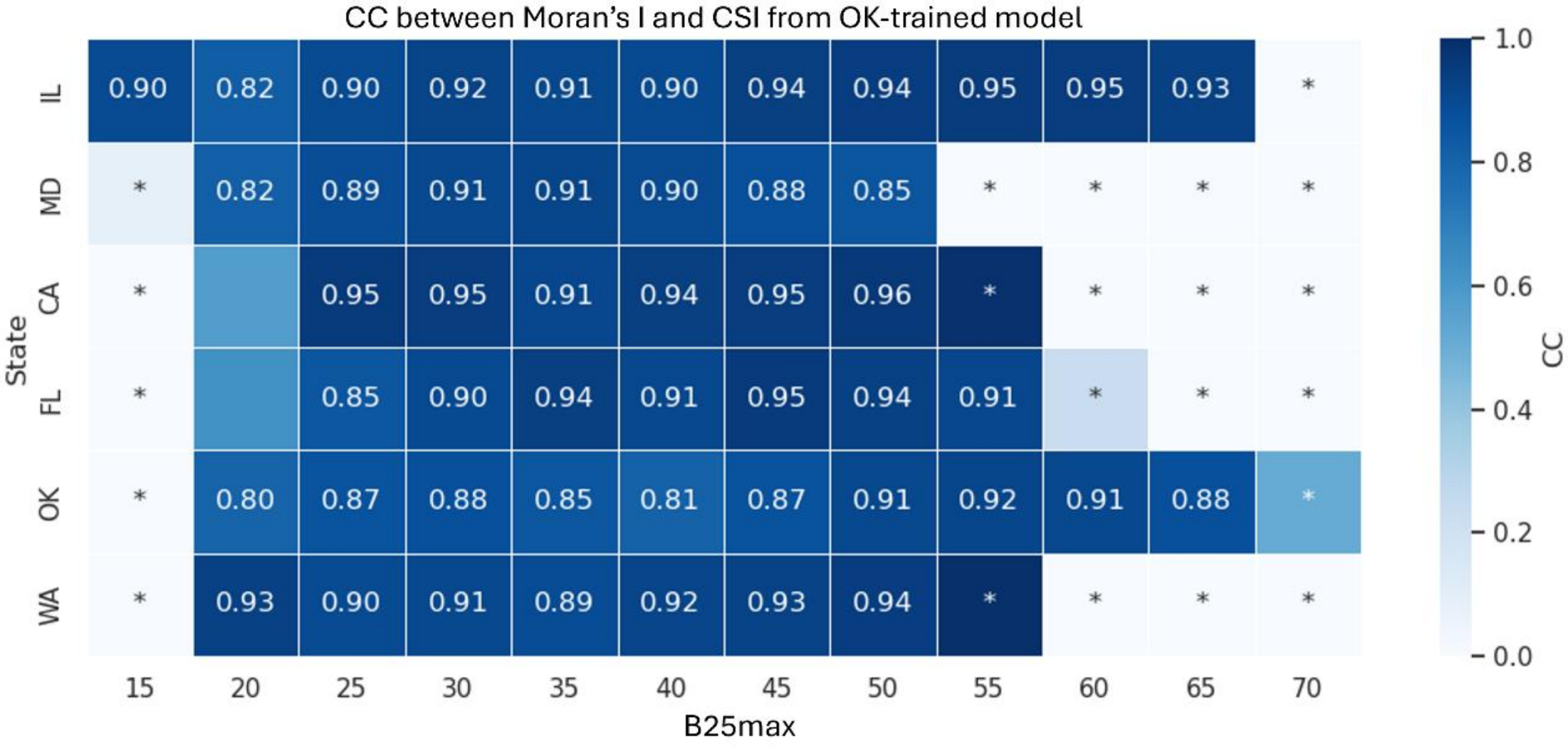

**Figure 7. Correlation between bin-specific Moran's I and CSI across regions and Bmax25 strata. Cell values are Pearson correlation coefficients; stars indicate strata with sample percentages below 1%.**

## 4.2 General metrics are necessary but insufficient for precipitation downscaling

General metrics remain necessary for evaluating precipitation downscaling, but the results show that they describe only one aspect of model behavior. PSNR, SSIM, $R^2$, and RMSE summarize full-field image fidelity and are useful for comparing average downscaling quality. However, full-field averaging can hide errors in localized high-reflectivity structures. In this study, this limitation is visible when image-domain, spectral, and distributional metric rankings diverge (Section 3.1). A downscaled field can achieve favorable image-domain scores while still missing or weakening small precipitation cores that are important for event detection.

This divergence is clear. It reflects a consistent difference between smooth and sharp reconstructions. The case studies in Figure 5 show that the ALL-trained output is generally smoother, whereas the OK-trained output retains sharper small-scale texture. Sharper fields better preserve the amplitude spectrum and the intensity distribution. This difference is visible even in OK, where neither model faces distribution shift. WDM-ALL achieves higher PSNR and lower RMSE, while WDM-OK keeps lower amplitude MAE and Wasserstein distance (Figure 2). Each metric family rewards a different reconstruction behavior, and no single family identifies the better model.

Bin-wise CSI complements these metric families by treating downscaling evaluation as a reflectivity-bin detection problem. Each reflectivity bin is evaluated separately rather than through exceedance above a single threshold (Section 2.4.2). CSI has a long history in rare-event and warning verification (Schaefer, 1990). In this study, the bin-wise CSI results show that Bicubic can remain competitive in low-reflectivity bins but loses skill in higher bins where localized structures are more important. This pattern would be obscured if evaluation relied only on PSNR, SSIM, $R^2$ or RMSE. However, bin-wise CSI yields one score per reflectivity bin, whereas model comparison and difficulty assessment ultimately require a sample-level summary.

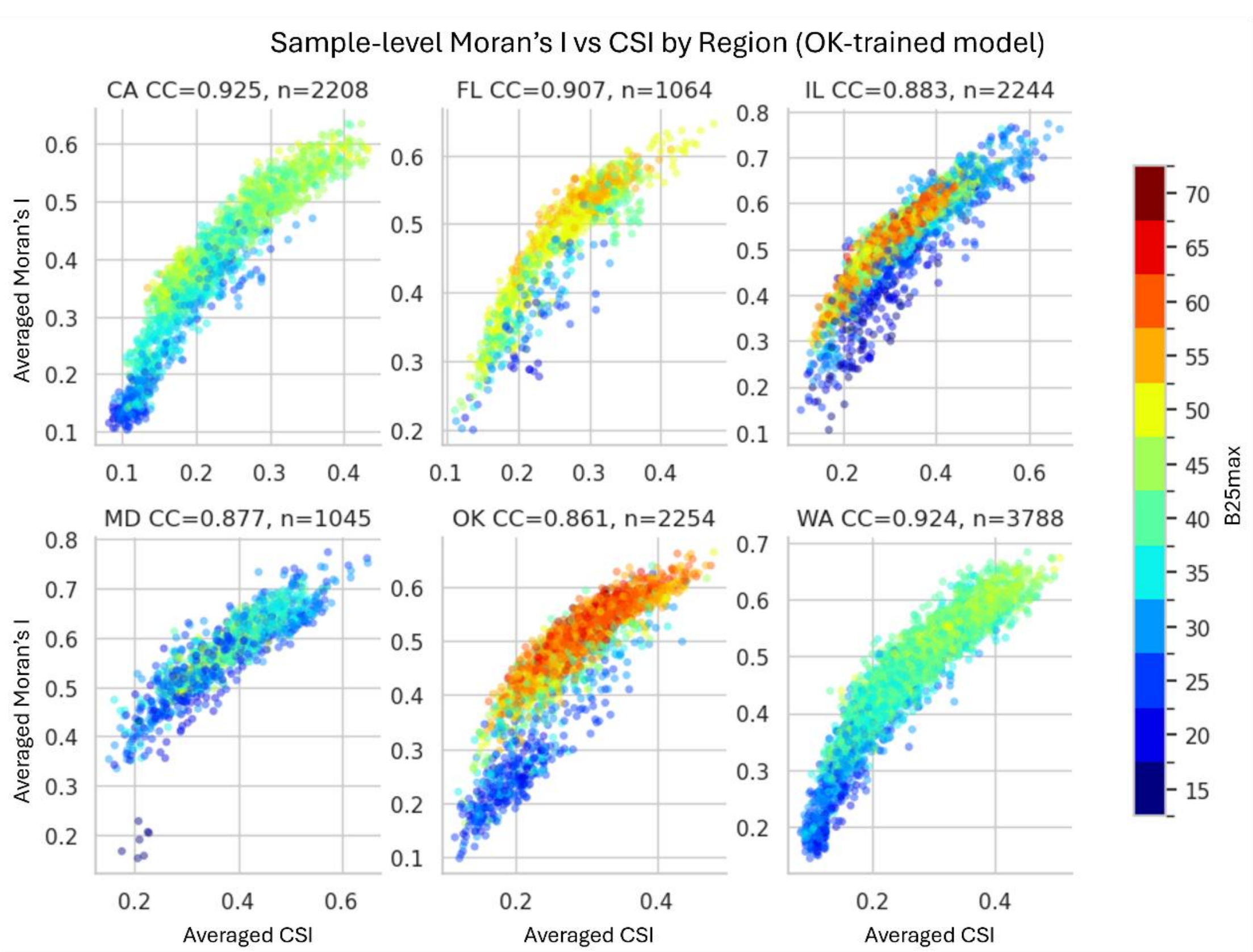

**Figure 8. Sample-level weighted Moran's I and weighted CSI by region. The CSI is calculated from WDM-OK. Moran's I and CSI are averaged over valid bins for each sample.**

To provide this summary, we average the bin-wise metrics over the valid reflectivity bins of each sample. This yields sample-level CSI and Moran's I. Averaged CSI describes the overall detection skill of a sample. Averaged Moran's I describes the average spatial autocorrelation of the valid reflectivity bins in the target field. Figure 8 compares the two quantities for WDM-OK in all six regions. The bin-wise relationship persists after this aggregation. The correlation pooled over all six regions is 0.901, and regional correlations range from 0.861 in Oklahoma to 0.925 in California. The five regions outside Oklahoma are out-of-distribution for this model, yet the correlation remains equally strong there. The link between spatial autocorrelation and detection skill therefore holds even under regional distribution shift. Bmax25 further shows that weak and intense samples follow the same relationship, so the correlation is unlikely to be an artifact of sample intensity. These results suggest that, for WDM-OK, sample-level detection skill varies systematically with the spatial autocorrelation of the target field. Whether this dependence is model-specific or fundamental to the downscaling task remains to be tested with other models. Together, bin-wise CSI and its sample-level average complement general image metrics with a spatially aware and intensity-aware view.

### 4.3 Regional robustness depends on the task boundary

Section 3 indicates that WDM achieves meaningful cross-region and cross-event generalization in this controlled setting, where the coarse inputs and fine targets are derived from the same MRMS product. WDM-ALL provides the most stable image-domain performance across regions, while WDM-OK retains competitive skill outside its source region and remains competitive in spectral and distributional fidelity. Together, these results indicate that part of the learned precipitation-structure prior is transferable across regions.

This generalization, however, must be interpreted within the boundary of the present task. In the same-product controlled setting, the coarse input and fine target come from the same observing system and share broadly consistent intensity distributions, occurrence characteristics and spatial-pattern statistics. Under this design, the model mainly needs to reconstruct fine-scale structure that is compatible with the coarse field. By contrast, in cross-product downscaling tasks, the source and target fields differ not only in resolution but also in their underlying statistical properties. For example, PrecipDiff enhances satellite precipitation estimates toward radar-like fields, which requires both spatial downscaling and correction of source-target product inconsistency (Dai & Ushijima-Mwesigwa, 2025). Similar combined correction-and-downscaling settings have also been explored for coarse global precipitation forecasts (Harris et al., 2022; Lopez-Gomez et al., 2025; Price & Rasp, 2022). Therefore, the cross-region generalization observed in this evaluation should not be assumed to transfer directly to cross-product correction-and-downscaling tasks, where the model must also learn systematic source-target mismatches, for example, the storm core area (Li et al., 2020). Within this controlled same-product setting, the Moran's I–CSI relationship becomes particularly important because it reveals how differences in spatial organization contribute to the sample variability in downscaling performance.

Besides the same-product setting, we should also pay attention to the choice of precipitation variables. To avoid uncertainties associated with quantitative precipitation estimation, we chose MRMS composite reflectivity to represent precipitation patterns. However, because rain rate depends quasi-exponentially on reflectivity in dBZ, averaging a reflectivity field over an area is not equivalent to averaging the corresponding precipitation field. Future work should couple downscaling with explicit bias correction for real low-resolution products, evaluate the probabilistic calibration of generated ensembles, and extend reflectivity-bin detection toward impact-relevant hydrological and warning metrics.

## 5 Conclusion

Generalization remains a central challenge for precipitation downscaling models because both the intensity distribution and the spatial organization of precipitation can vary substantially across climate and storm regimes. This study examined whether a WDM trained in one region can retain useful downscaling ability in other regions under a controlled same-product MRMS reflectivity super-resolution setting. To isolate the spatial super-resolution problem from product-to-product bias correction, the low-resolution inputs and high-resolution targets were derived from the same MRMS fields. We compared a WDM-OK and a WDM-ALL against NN and Bicubic interpolation across six U.S. regions, using image-domain reconstruction metrics, spectral and distributional metrics, bin-wise CSI, case studies and Moran's I-based spatial-autocorrelation analysis.

The main conclusions are as follows:

- WDM shows meaningful cross-region and cross-event generalization in the same-product MRMS downscaling setting. Even when trained only on Oklahoma cases, WDM preserves much of its downscaling skill in the five geographically unseen regions, indicating that part of the learned precipitation-structure prior transfers across regional settings. WDM-ALL generally provides more stable and smoother overall performance, increasing PSNR by 1.14-1.79 dB and reducing RMSE by 12.6-18.0% relative to NN across the six test regions.
- Bin-wise evaluation reveals how regional generalization differs across precipitation intensities. Image-domain metrics provide useful information about average reconstruction fidelity, but they do not fully show whether a model transfers skill for specific reflectivity ranges. Bin-wise CSI fills this gap by evaluating non-overlapping reflectivity bins separately. Low-reflectivity structures can remain relatively easy for interpolation, whereas higher-reflectivity structures more clearly expose the differences between WDM and the baselines.
- Spatial autocorrelation provides a useful diagnostic of downscaling performance by helping explain why some precipitation structures are reconstructed more successfully than others despite regional differences. The large CSI variability within individual reflectivity bins is not explained by intensity alone. Moran's I analysis shows that bins with stronger spatial autocorrelation are generally more detectable after downscaling, whereas fragmented structures are more difficult to reconstruct. This relationship persists across regions and precipitation-intensity strata. For WDM-OK, the weighted Moran's I–CSI correlation is 0.901 over all regions, and regional values range from 0.861 in Oklahoma to 0.925 in California.

Together, these findings show that cross-region generalization in precipitation downscaling cannot be characterized solely by whether the target region appears in training. In a controlled same-product downscaling setting, WDM can learn useful precipitation-structure information across regions, but the apparent strength of that generalization depends on the spatial organization of the target field. The bin-wise metrics and diagnostics developed here therefore help distinguish cross-region generalization, intensity-specific reconstruction skill, and morphology-dependent difficulty. Ultimately, precipitation downscaling should be evaluated not only by improvements in average image similarity, but also by the failure modes that determine whether event-relevant structures are preserved.

## Data and code availability

NOAA Multi-Radar/Multi-Sensor System (MRMS) is publicly accessible from https://registry.opendata.aws/noaa-mrms-pds. The Storm Event Database from NOAA National Centers for Environmental Information is publicly accessible from https://www.ncei.noaa.gov/stormevents/. WDM model code is available in Zenodo (Yi et al., 2025).

# Supporting Information

## Contents of this file

Text S1

Tables S1 to S2

Figures S1

## Introduction

This supporting information provides supplementary methodological details, tabulated dataset information and additional model-training results for the evaluation of regional robustness and spatial-structure performance in wavelet-diffusion precipitation downscaling.

## Text

### Text S1

This section provides the mathematical definitions of metrics used in this study. Let $y$ denote the target high-resolution field, $\hat{y}$ denote the predicted field and $N$ denote the number of pixels. All metrics were computed after applying the same preprocessing, masking and normalization settings used in the main evaluation.

The root mean squared error (RMSE) between the prediction and target is

$$RMSE = \sqrt{\frac{1}{N}\sum_{i=1}^{N}(\hat{y}_i - y_i)^2}$$

Peak Signal-to-Noise Ratio (PSNR) is computed by

$$PSNR = 10\,log_{10}\left(\frac{MAX(y)}{\frac{1}{H*W}\sum_{i=1}^{H}\sum_{j=1}^{W}\left(y_{ij} - \hat{y}_{ij}\right)^2}\right)$$

Structural Similarity Index Measure (SSIM) is defined as

$$SSIM(x,y) = \frac{2\mu_x\mu_y + C_1}{\mu_x^2 + \mu_y^2 + C_1} * \frac{2\sigma_x\sigma_y + C_2}{\sigma_x^2 + \sigma_y^2 + C_2} * \frac{\sigma_{xy} + C_3}{\sigma_x\sigma_y + C_3}$$

R squared measures how well observed outcomes are replicated by the model, which is defined as

$$R^2 = 1 - \frac{\sum_i (y_i - \hat{y}_i)^2}{\sum_i (y_i - \bar{y}_i)^2}$$

The amplitude MAE is calculated on frequency domain transferred from image-domain by fast Fourier transform, which is calculated by

$$AmpMAE = \frac{1}{HW}\sum_{i=0}^{H-1}\sum_{j=0}^{W-1}\left|\left|\mathcal{F}_y(i,j)\right| - \left|\mathcal{F}_{\hat{y}}(i,j)\right|\right|$$

Where $\mathcal{F}_x(u,v) = \frac{1}{HW}\sum_{n=0}^{H-1}\sum_{m=0}^{W-1} x(n,m)\exp(-2\pi i\left(\frac{um}{H} + \frac{vn}{W}\right))$.

One-dimensional Wasserstein distance is canulated when optimal transport preserves the order of probability mass elements, which is defined by

$$W_1(y,\hat{y}\ ) = \int \left|F_y(x) - F_{\hat{y}}(x)\right| dx$$

Where $F_y$ and $F_{\hat{y}}$ are the cumulative distribution functions (CDF) of $y$ and $\hat{y}$.

## Tables

**Table S 1. Study regions, crop windows and event-selection rules used to construct event-centered MRMS samples.**

| Region | Longitude window | Latitude window | Precipitation type |
|---|---|---|---|
| OK | [-99, -96] | [34.5, 37.5] | Strong convective events: Hail, Tornado, Thunderstorm Wind and High Wind |
| IL | [-88.5, -85.5] | [40, 43] | Winter precipitation, hail and heavy-precipitation-related events |
| WA | [-123.75, -120.75] | [46, 49] | Heavy Snow |
| FL | [-83, -80] | [26, 29] | Tropical Storm and Tropical Depression events |
| MD | [-78.5, -75.5] | [37, 40] | Winter Weather and Ice Storm |
| CA | [-123, -120] | [36, 39] | Records with the narrative keyword "atmospheric river" |

**Table S 2. Sample counts by region and year. Train and validation counts are approximate because data in 2022-2023 is randomly split at an 80/20 ratio. Each sample is one 10-min MRMS time interval.**

| Region | 2022 | 2023 | Train/Val total | Train | Val | 2024 test |
|---|---|---|---|---|---|---|
| OK | 1732 | 2307 | 4039 | 3231 | 808 | 2254 |
| IL | 2484 | 1719 | 4203 | 3362 | 841 | 2244 |
| WA | 5216 | 4035 | 9251 | 7400 | 1851 | 3788 |
| FL | 1003 | 235 | 1238 | 990 | 248 | 1064 |

| MD | 1835 | 1482 | 3317 | 2653 | 664 | 1045 |
|---|---|---|---|---|---|---|
| CA | 1026 | 4110 | 5136 | 4108 | 1028 | 2208 |
| **Total** | **13296** | **13888** | **27184** | **21747** | **5437** | **12603** |

## Figures

Figure S 1 compares the original-field and residual-field parameterizations on both training splits. The diffusion-loss and MSE panels are reported in each model's own target space and thus differ in scale by construction; the comparison rests on the validation-sample MSE, computed on sampled outputs in dBZ space (bottom panel). Two observations support the choice made in the main text. First, the residual-field models converge substantially faster: their best checkpoints, selected by minimum validation-sample MSE, occur at epoch 460 (ALL-trained) and epoch 1220 (OK-trianed), whereas the original-field models require epochs 1900 and 2980, respectively, to reach their optima — roughly a three- to four-fold difference. This is also visible in the validation-sample MSE curves, where the residual models reach their plateau well before epoch 1000. Second, the converged performance is comparable: the best validation-sample MSE is 7.71 vs. 7.63 (all-region) and 8.32 vs. 8.02 (random-split) for the residual and original variants, a difference of 1–4%. Given the comparable validation accuracy and markedly faster convergence, the residual-field models were selected for the main evaluation.

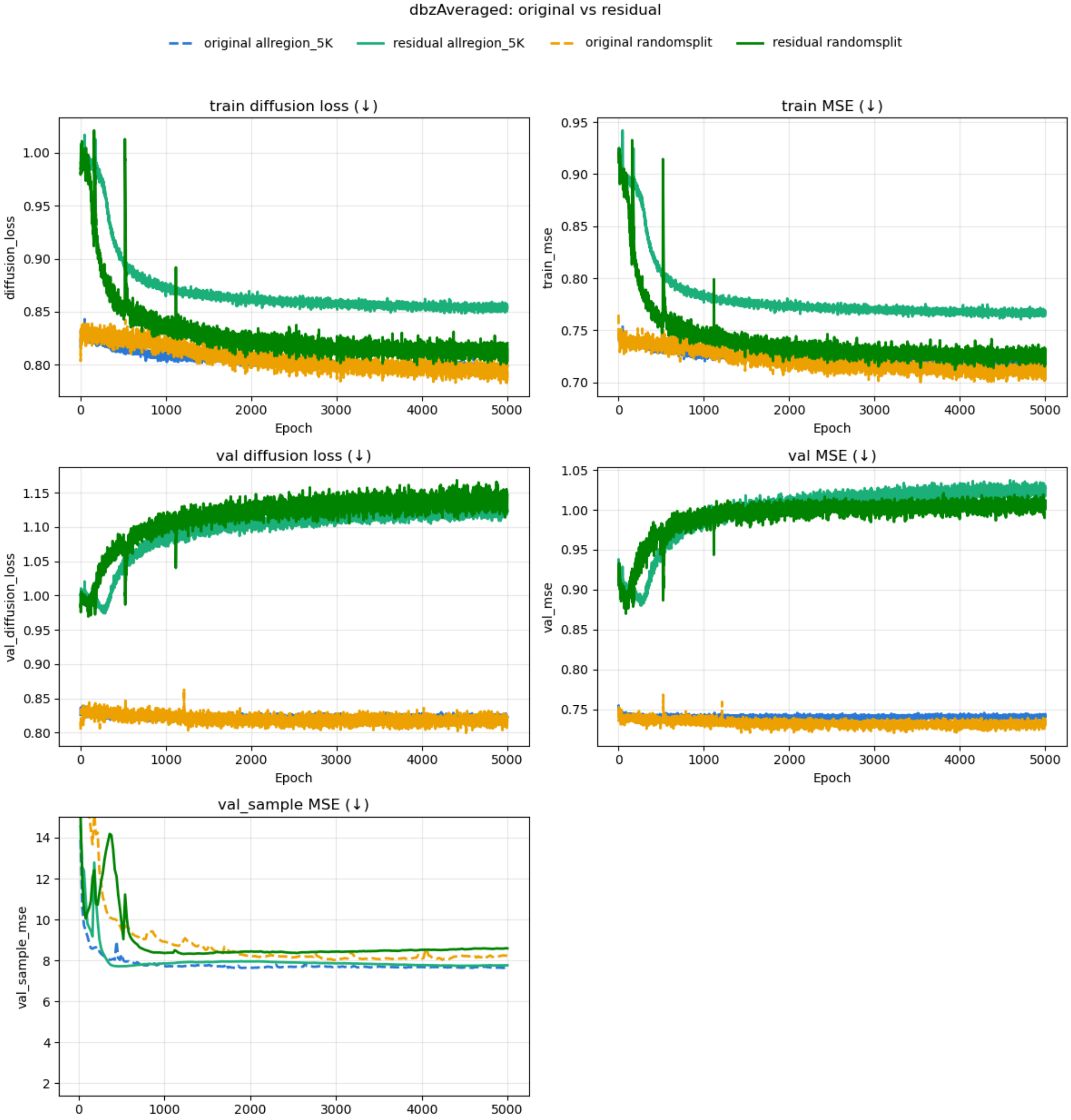

**Figure S 1 Training and validation histories of the original-field (dashed) and residual-field (solid) WDM variants under the dbzAveraged configuration, for the all-region (5K-epoch) and random-split (OK-region) training sets. Top and middle rows: per-epoch diffusion loss and MSE on the training and validation sets; these are computed in each parameterization's own target space (full field vs. residual field) and are therefore not directly comparable across the two variants. Bottom: MSE of sampled outputs on the validation set, computed in dBZ space and directly comparable across variants (y-axis truncated at 15 to crop the initial transient).**